\pdfoutput=1

\documentclass[aps,prx,reprint,superscriptaddress]{revtex4-2}
\usepackage{graphicx}  
\usepackage{amsmath,amssymb}   
\usepackage{hyperref}  

\usepackage{color}

\begin{document}

\title{Complex fractal trainability boundary can arise from trivial non-convexity}
\author{Yizhou Liu}
\email{liuyz@mit.edu}
\affiliation{
 Physics of Living Systems, Department of Physics, MIT, Cambridge, MA 02139, USA.
}
\affiliation{
 Department of Mechanical Engineering, MIT, Cambridge, MA 02139, USA.
}

\date{\today}

\begin{abstract}
Training neural networks involves optimizing parameters to minimize a loss function, where the nature of the loss function and the optimization strategy are crucial for effective training. Hyperparameter choices, such as the learning rate in gradient descent (GD), significantly affect the success and speed of convergence. Recent studies indicate that the boundary between bounded and divergent hyperparameters can be fractal, complicating reliable hyperparameter selection. However, the nature of this fractal boundary and methods to avoid it remain unclear. In this study, we focus on GD to investigate the loss landscape properties that might lead to fractal trainability boundaries. We discovered that fractal boundaries can emerge from simple non-convex perturbations, i.e., adding or multiplying cosine type perturbations to quadratic functions. The observed fractal dimensions are influenced by factors like parameter dimension, type of non-convexity, perturbation wavelength, and perturbation amplitude. Our analysis identifies ``roughness of perturbation", which measures the gradient's sensitivity to parameter changes, as the factor controlling fractal dimensions of trainability boundaries. We observed a clear transition from non-fractal to fractal trainability boundaries as roughness increases, with the critical roughness causing the perturbed loss function non-convex. Thus, we conclude that fractal trainability boundaries can arise from very simple non-convexity. We anticipate that our findings will enhance the understanding of complex behaviors during neural network training, leading to more consistent and predictable training strategies.
\end{abstract}

\maketitle

\section*{Introduction}
Machine learning has become a cornerstone of modern technology. 
When training a machine learning model,
we need to update the model parameters towards optimizing a loss function (usually based on the loss gradient) to attain desired performance.
To better understand and achieve successful training, researchers have tried to capture shapes of the loss landscapes \cite{choromanska15,Jin2018OnTL,zhang2022symmetry} and dynamics that arise from optimization algorithms \cite{SuBoydCandes16,Wibisono16, KongTao20, ShiSuJordan23, chen2023stability}.
In general, despite the considerable empirical success and broad application of these optimization techniques in training models, our theoretical understanding of the training processes remains limited.

Recent empirical evidence suggests that whether a model is trainable can be extremely sensitive to choices in optimization.
A model is said to be not trainable here if the optimization applied leads to divergent loss function value.
By convention, we call model parameters to be optimized as parameters and other parameters in the optimizer controlling the optimization process as hyperparameters. One of the most important hyperparameters is learning rate, which affects the size of the steps taken during optimization.
On a simple two layer neural network, gradient descent (GD) was recently found to have a fractal boundary between learning rates that lead to bounded and divergent loss (fractal trainability boundary for short) \cite{sohldickstein2024boundary}. Consequently, a slight change in hyperparameters can change the training result qualitatively with little hope to choose good hyperparameters in advance.

Here, we aim to explore the mechanisms underlying the fractal trainability boundary and the key factors influencing its fractal dimension. Guided by the intuition that non-convexity renders the gradient sensitive to parameter, which makes GD sensitive to varying learning rates, we tried to quantify the relation between non-convexity and fractal trainability. Given the difficulties in describing and controlling the loss functions of real neural networks, our approach involves constructing simple non-convex loss functions and testing GD on these to examine the boundary between learning rates that lead to bounded versus divergent losses. We discover that even a simple quadratic function, when perturbed by adding or multiplying a regular perturbation function (specifically a cosine function), exhibits a fractal trainability boundary. A parameter specific to the form of the perturbation, defined as roughness, appears to govern the fractal dimension of the trainability boundary, describing the gradient sensitivity to the parameter. A notable difference emerges between the fractal behaviors of trainability in quadratic functions perturbed by additive perturbation versus those altered by multiplicative perturbation: fractal behavior disappears at a finite roughness (when the perturbed loss becomes convex) for additive cases but persists for multiplicative perturbations (where the perturbed loss is always non-convex). We therefore offer a perspective to explain the fractal trainability boundaries observed in real neural networks as a result of non-convexity, emphasizing ``roughness" as the factor controlling fractal dimensions.

\section*{Results}

We first elaborate the key intuition of why certain non-convexity may lead to the fractal trainability boundary. 
Common ways to generate fractals involve iterating a function sensitive to hyperparameters in the function (e.g., Mandelbrot and quadratic Julia sets \cite{mandelbrot1982fractal, sohldickstein2024boundary}).
We denote the loss function as $f(x)$ where $x$ is the parameter to optimize to minimize $f(x)$. GD can be described as iterating the function
\begin{equation}
	x^{(k+1)} = x^{(k+1)}(x^{(k)}; s) = x^{(k)} - s \nabla f(x^{(k)} ).
\end{equation}
Here, $x^{(k)}$ is the parameter obtained at the $k$th step and $s$ is the learning rate (hyperparameter).
If non-convexity can make the gradient $\nabla f(x)$ sensitive to parameter $x$, after we shift learning rate $s$ a little at the $k$th step (this is another training process to be compared with the original one), $x^{(k+1)}$ will be a little different from the one obtained with the unchanged learning rate. However, the gradient at the new $x^{(k+1)}$ will be very different, leading to very different subsequent iterations. The sensitivity of gradient's dependence on parameter can therefore be transformed to the sensitive dependence of optimization process on the learning rate (hyperparameter), which is the key to generate fractals. Notably, this sensitive dependence on learning rate is sufficient to generate chaos while is not obvious to yield divergent training.
\begin{figure}
    \centering
    \includegraphics{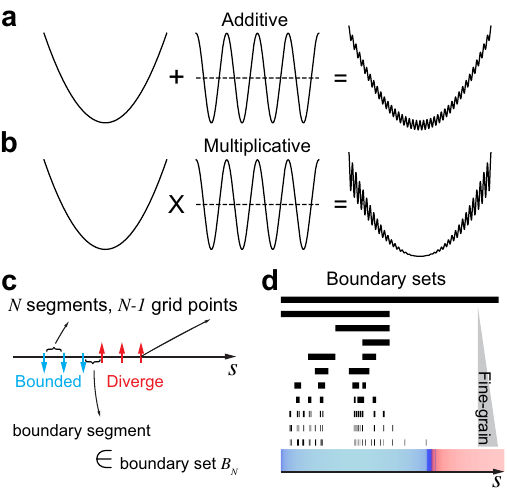}
    \caption{On constructed loss functions, we conduct numerical experiments to study the trainability boundaries. (a) Illustration of loss landscapes with additive perturbation ($f_+$ with $\epsilon = 0.2$ and $\lambda = 0.1$). (b) An example of loss constructed having multiplicative perturbation ($f_\times$ with $\epsilon = 0.2$ and $\lambda = 0.1$). (c) On a fixed range of learning rate, we can put in $N$ small segments and evaluate whether training diverge or not at each end of the segments. We therefore can generate a set of boundary segments, $B_N$ and count the number of boundary segments. (d) An example when we have more segments (fine-grain), the number of boundary segments (black segments) is increasing (figure obtained based on multiplicative perturbation case $f_\times$ with parameters $\epsilon = 0.2$ and $\lambda = 0.1$). The colored bar at the bottom visualizes losses for bounded (blue) and divergent training (red).}
    \label{fig:1}
\end{figure}

We thus need to do experiments on specific functions to test if non-convexity with sensitive gradients can lead to fractal trainability boundaries. 
We started by looking at one dimensional parameters ($x\in \mathbb{R}^d$, $d=1$) and chose to construct our loss landscape by perturbing the convex quadratic function $f_0(x) = x^2$. One way of perturbation is to add a non-convex function (Fig.~\ref{fig:1}a). We used the simplest regular perturbation $\epsilon f_1 =\epsilon \cos(2\pi x/\lambda)$, where $\lambda$ is the wavelength of the perturbation, and $\epsilon$ the amplitude of the perturbation. We therefore defined the additive perturbation case in our context as
\begin{equation}
f_+(x) = f_0 + \epsilon f_1 = x^2 + \epsilon \cos(2\pi x/\lambda).
\end{equation}
An alternative way to introduce non-convexity is via multiplying a perturbation function (Fig.~\ref{fig:1}b), which will be referred to as multiplicative perturbation case:
\begin{equation}
f_\times (x) = f_0 (1 + \epsilon f_1) = x^2 (1 + \epsilon \cos(2\pi x/\lambda)).
\end{equation}
The two cases are qualitatively different as when $x\to \infty$, the additive perturbation will become small comparing to $f_0 = x^2$ while the multiplicative perturbation is always comparable to the unperturbed $f_0 = x^2$.
Our test functions have simple analytic forms and represent different forms of non-convexity.

We next explain the idea of investigating fractal trainability boundaries numerically. The boundary points separating learning rates leading to finite and divergent loss are not accessible directly. So, we need to use finite but many grid points to locate the boundary learning rates. On a given range of learning rate ($s \in [s_{\rm min}, s_{\rm max}]$), we can evenly put $N+1$ grid points, with which GD can be tested to diverge or not. If two neighboring learning rate grid point values lead to the same divergent/bounded loss behavior, at this coarse-grain level (quantified by $N$), we say there is no boundary between the two grid points. Otherwise, we say the segment between this two grid points is a boundary segment. We define a set $B_N$ as the set containing all boundary segments when we have $N+1$ grid points (Fig.~\ref{fig:1}c). Heuristically, we expect each boundary segment to cover one boundary accurately when the segment length goes to zero (i.e., $N\to \infty$), and thus $B_N$ becomes the set of boundary learning rates when $N\to \infty$. If the number of boundary segments, denoted by $|B_N|$, increases with respect to $N$ as a scaling law asymptotically
\begin{equation}
 |B_N| \propto N^\alpha,~(N\to \infty),
\end{equation}
we say the trainability boundary has a fractal dimension $\alpha$ (by convention, the fractal dimension defined in this way is called box dimension \cite{strogatz2018nonlinear}). We observed that the number of boundary segments (black segments in Fig.~\ref{fig:1}d) indeed increases when we do tests on the multiplicative case and put more and more testing grid points in a fixed learning rate range (Fig.~\ref{fig:1}d). The colored bar at the bottom visualizes loss values for bounded (blue) and divergent (red) training evaluated at $2^{20}$ grid points in $[0,1.5]$ (Fig.~\ref{fig:1}d). And the color intensity is determined by $\sum_i f_i$ for bounded training and $\sum_i f_i^{-1}$ for divergent training, respectively \cite{sohldickstein2024boundary}, where $f_i$ is the loss value at $i$th step (totally 1000 steps). Once we zoom in, we see more boundaries between blue and red (the original vector image can be found \href{https://github.com/liuyz0/FractalBoundary/tree/main/figures}{online}).
We therefore are ready for further exploration with the numerical tool identifying fractals.

\begin{figure}
    \centering
    \includegraphics{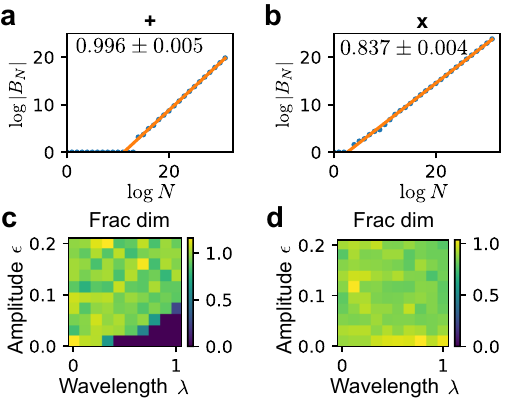}
    \caption{Simple non-convexity can lead to fractal trainability boundaries, whose fractal dimensions depend on perturbation form,
wavelength, and amplitude. (a) For additive perturbation case $f_+$ with $\epsilon = 0.2$ and $\lambda = 0.1$, we studied learning rate in $[0,1.5]$, where the number of boundary segments increases as a scaling law with respect to the number of segments put, suggesting a fractal trainability boundary. The fractal dimension, i.e., the slope $\log|B_N|$ against $\log N$ is fitted as $0.996 \pm 0.005$. (b) For additive perturbation case $f_\times$ with $\epsilon = 0.2$ and $\lambda = 0.1$, fractal trainability boundary is also observed with fractal dimension $0.837 \pm 0.004$. (c, d) The fractal dimension of trainability boundary vary with respect to perturbation amplitude $\epsilon$ and wavelength $\lambda$. In particular, for the additive perturbation case (c), the fractal dimension increases with larger amplitude and smaller wavelength. For the multiplicative case (d), the fractal dimension has no clear dependence on the two function parameters.}
    \label{fig:2}
\end{figure}
We investigated the trainability boundaries using GD on specifically constructed loss functions. For our experiments, we applied GD to the loss function with additive perturbation, $ f_+ $, starting from $ x^{(0)} = 1.0 $ with parameters $ \epsilon = 0.2 $ and $ \lambda = 0.1 $. We conducted 1000 steps of GD and defined a training session as divergent (untrainable) if the sum of losses at the 1000 steps exceeded $ 10^16 $. This upper loss threshold affects misclassifying slowly diverging training into bounded training; although varying it between $ 10^12 $ and $ 10^{20} $ did not alter the observed fractal dimension. We adjusted the learning rate within the range of $[0, 1.5]$ and increased the number of intervals, $ N $, up to $ 2^{32} $ (see Methods). Our findings reveal that the trainability boundary for this simple function displays fractal behaviors, meaning the number of boundary segments, $ |B_N| $, increases following a scaling law with $ N $ at large values (Fig.~\ref{fig:2}a). The fractal dimension, $ \alpha $, calculated via least squares as the slope of $ \log |B_N| $ against $ \log N $ ($\log$ base is $2$ throughout this paper), is approximately $ 0.996 \pm 0.005$ (error is standard deviation). A similar analysis on another loss function with multiplicative perturbation, $ f_\times $, yielded a fractal dimension of $ \alpha = 0.837 \pm 0.004$ (Fig.~\ref{fig:2}b). This suggests that fractal boundaries are more densely packed in a narrower range for the additive perturbation scenario, indicating a potentially less erratic behavior. Nonetheless, the emergence of fractal trainability boundaries in these trivially simple loss functions is remarkable.

We next sought to examine factors affecting the fractal dimension of trainability boundaries. 
We evaluated the fractal dimension of the trainability boundary with ten different amplitudes $\epsilon$ evenly picked from $[0.01, 0.2]$ and ten different wavelengths $\lambda$ evenly picked from $[0.01, 1.0]$. Least square fitting is used to obtain the fractal dimensions. We found that for the additive perturbation case, the fractal dimension increases with decreasing wavelength and increasing amplitude (Fig.~\ref{fig:2}c), while for the multiplicative perturbation case, the fractal dimension has no clear dependence on amplitude or wavelength (Fig.~\ref{fig:2}d). The fractal dimension therefore depends on the type, wavelength, and amplitude of the non-convex perturbation in a complicated manner.

We next tried to analyze how the perturbation wavelength and amplitude change the fractal dimension of trainability boundary. We can rescale the parameter, $\tilde{x} = x/b$, and renormalize the loss, $\tilde{f}(\tilde{x}) = f(x) / \zeta$, such that we map a GD process starting at $x^{(0)}$ to another one starting at $\tilde{x}^{(0)} = x^{(0)}/b$ and updating with respect to
\begin{equation}
\tilde{x}^{(k+1)} = \tilde{x}^{(k)} - \tilde{s} \nabla \tilde{f}(\tilde{x}^{(k)}).
\end{equation}
By choosing $\zeta = b^2$, we can show that $\tilde{s} = s$ and $\tilde{f}$ has the same function form (i.e., with additive perturbation or multiplicative perturbation) as $f$ while with different function parameters $\tilde{\epsilon}$ and $\tilde{\lambda}$ (see Methods). This means a GD process on our constructed loss given $\epsilon$, $\lambda$, and $x^{(0)}$ will have the same divergence property with the same learning rate as another GD process on our constructed loss with $\tilde{\epsilon}$, $\tilde{\lambda}$, and $\tilde{x}^{(0)}$. Therefore, trainability boundaries are the same for the two sets of conditions, $\{\epsilon, \lambda, x^{(0)}\}$ and $\{\tilde{\epsilon}, \tilde{\lambda}, \tilde{x}^{(0)}\}$. If we further assume the fractal dimension $\alpha$ depends mainly on loss properties rather than initial conditions (seems to be true, see SI), we conclude that fractal dimensions of the trainability boundaries for $\{\epsilon, \lambda\}$ and $\{\tilde{\epsilon}, \tilde{\lambda}\}$ are the same. 

\begin{figure}
    \centering
    \includegraphics{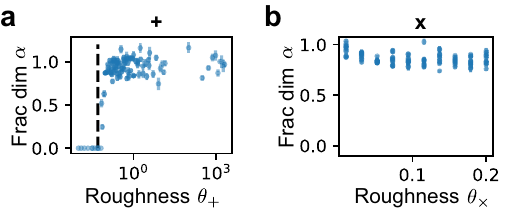}
    \caption{Roughness determines fractal dimension of trainability boundaries and captures the transition to fractal trainability boundary when the landscape is non-convex. (a) For the additive perturbation case, the roughness $\theta_+$ found well organizes the fractal dimensions $\alpha$ with different amplitude and wavelength (data in Fig.~\ref{fig:2}c), where we can see a clear transition to non-zero fractal dimensions near $\theta_+ = 1/2\pi^2$ (dashed line, corresponding emergence of non-convexity). (b) For the multiplicative perturbation case, the roughness $\theta_\times$ found determines the fractal dimensions $\alpha$ (data from Fig.~\ref{fig:2}d). Error bars are standard deviations of fitting.}
    \label{fig:3}
\end{figure}

More specifically, for the additive perturbation case, the renormalization flow not changing the fractal dimension should be given by (see Methods)
\begin{equation}
\tilde{\epsilon} = \epsilon/b^2,~\tilde{\lambda} = \lambda/b.
\end{equation}
Note there is only one independent combination of $\epsilon$ and $\lambda$, i.e.,
\begin{equation}
\theta_+ = \epsilon/\lambda^2,
\end{equation}
is invariant under the renormalization transformation, we claim the fractal dimension can only depends on $\theta_+$ since it only depends on $\{\epsilon, \lambda\}$ and is invariant under the renormalization transformations.
We would call the quantity $\theta_+$ ``roughness" as it is in the pre-factor of the second derivative of the additive perturbation, measuring the sensitivity of gradient's dependence on parameter. We plotted the fractal dimension as a function of roughness $\theta_+$ (the same set of data as Fig.~\ref{fig:2}c) and found the fractal dimension shows a clear and sharp transition from zero (i.e., no fractal behavior) to non-zero when increasing roughness $\theta_+$ (Fig.~\ref{fig:3}a). We found that the critical roughness $\theta_+$ is near $1/2\pi^2$ (dashed line in Fig.~\ref{fig:3}a), which corresponds to the critical situation $f_+$ begin to be non-convex ($\exists x$ such that $\nabla^2 f_+(x) = 0$). The simple renormalization analysis yields roughness of the additive perturbation, which determines the fractal dimension and shows that fractal behaviors show up when the perturbed loss is non-convex.

Following the same renormalization procedure, we found the roughness determining fractal dimension for the multiplicative perturbation case is
\begin{equation}
\theta_\times = \epsilon.
\end{equation}
This quantity $\theta_\times = \epsilon$ also shows up in the second derivative of $f_\times$ and contributes to the sensitive dependence of gradient on parameter $x$, while it is not the only one (there are also $\epsilon/\lambda$ and $\epsilon/\lambda^2$) and thus just looking at the second derivative does not suffice. For the multiplicative perturbation case, we found the fractal dimension decreases a little with increasing roughness (Fig.~\ref{fig:3}b; same data as Fig.~\ref{fig:2}d).
The multiplicative case is always non-convex and always has fractal trainability boundary, which is consistent with the previous finding that fractal behaviors emerges when the perturbed loss becomes non-convex. Note that the non-convex part of the multiplicative case (where the second derivatives begin to be non-positive) has loss value around and above the order of magnitude $1/\epsilon$, if we have too small $\epsilon$ while our numerical upper bound to classify bounded and divergent training is not large enough, the classification will solely depends on the convex part of the loss. In this case, if non-convexity is necessary for fractal behaviors, we would expect no fractal behaviors as a numerical artifact, which is proved to be true (see SI). 
We therefore conclude that roughness found through simple renormalization determines fractal dimension of trainability boundaries and the transition to non-zero fractal dimensions corresponds to the loss function becoming non-convex.

\begin{figure}
    \centering
    \includegraphics{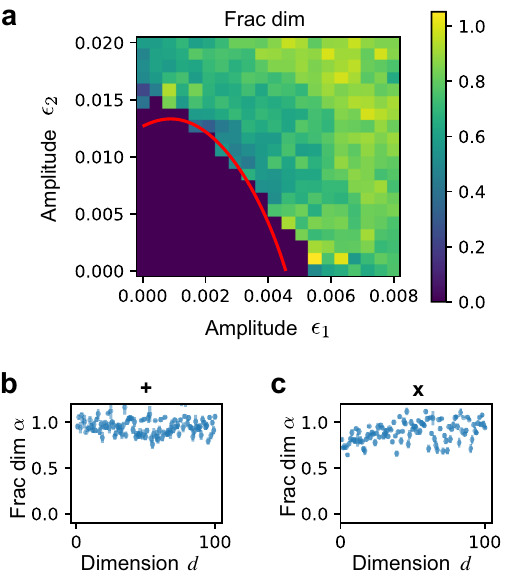}
    \caption{Beyond simple cases we can analyze, fractal dimension of trainability boundary depends on many other parameters determine the loss function, while it seems to be general that non-convexity leads to fractal behaviors.
    (a) For additive case with two cosine perturbations, the fractal dimension depends complicatedly on the amplitudes, while it is true fractal behaviors show up after the loss is non-convex (red line is the boundary of convexity).
    (b and c) For high dimensional optimization, the fractal dimension can depend on parameter dimensions. The fractal dimension is robust to increasing the parameter dimension $d$ for the additive perturbation case. (b) While the fractal dimension increases with the parameter dimension $d$ for the multiplicative perturbation case. Error bars are standard deviations of fitting.}
    \label{fig:4}
\end{figure}

Beyond simple cases we can analyze, we next studied a slightly more complicated loss landscapes. As a first step towards perturbations with multiple length scales, we considered additive perturbations with two cosine functions,
\begin{equation}
f_+(x) = x^2 + \epsilon_1 \cos(2\pi x/\lambda_1) + \epsilon_2 \cos(2\pi x/\lambda_2).
\end{equation}
The renormalization can only say $\epsilon_1/\lambda_1^2$ and $\epsilon_2/\lambda_2^2$ determine the fractal dimension and cannot yield a single variable controlling the fractal behavior.
In numerical tests, we fixed $\lambda_1 = 0.3$ and $\lambda_1 = 0.5$, while changed amplitudes $\epsilon_1$ and $\epsilon_2$. We found the fractal dimension depends non-monotonically on each of the amplitudes (Fig.~\ref{fig:4}a). However, the claim that fractal behaviors arise when the loss becomes non-convex is still valid, 
as the boundary between convex and non-convex losses (red curve in Fig.~\ref{fig:4}a, solved numerically) also separates zero and non-zero fractal dimensions.

We next explored how the dimension of parameters $x$ affect the trainability boundary.
For the additive perturbation case, we generalize the function for $x\in \mathbb{R}^d$ as
\begin{equation}
f_+(x) = \sum_i x_i^2 + \epsilon \sum_i \cos(2\pi x_i/\lambda).
\end{equation}
Following similar numerical methods as before (see detailed parameter setting in SI), we studied the fractal dimension of trainability boundary for $f_+$ varying $d$ from $1$ to $100$. The results suggest that the fractal dimension does not change much with respect to $d$ in the additive perturbation scenario (Fig.~\ref{fig:4}b). For the multiplicative perturbation case, we can define a class of functions
\begin{equation}
f_\times (x) =  \Big(1+ \epsilon \sum_i \cos(2\pi x_i/\lambda)\Big) \sum_i x_i^2.
\end{equation}
We found the fractal dimension $\alpha$ slowly increases in this case with respect to increasing $d$ (Fig.~\ref{fig:4}c), which makes sense as high-dimensional optimization should be more complicated. Our renormalization procedure cannot connect two functions with different dimensions $d$, and therefore roughness values for functions with different dimensions $d$ are not comparable. Future works are needed to analyze the impact of parameter dimensions $d$, e.g., defining a generalized roughness that can determine fractal dimension of trainability boundaries across different $d$. 

\section*{Discussion}

In this study, we have demonstrated that fractal trainability boundaries can arise from relatively simple non-convex modifications to loss functions. Specifically, our results show that the sensitivity of the loss gradient to parameter changes—a consequence of non-convexity introduced either through additive or multiplicative cosine perturbation—plays a crucial role in the emergence of fractal trainability. The fractal dimensions we observed are influenced by several factors, including the parameter dimension, the type of non-convexity, perturbation wavelength, and amplitude. Notably, our use of renormalization techniques in one-dimensional optimization cases has linked various loss functions to corresponding fractal dimensions of their trainability boundaries. Therefore, we have identified ``roughness of perturbation" as a key property that quantifies this sensitivity and dictates the fractal behavior. We observed a clear transition from non-fractal to fractal trainability boundaries as roughness increases, with the critical roughness causing the perturbed loss to be non-convex. These findings not only validate our hypothesis about the impact of non-convexity on trainability but also open up new avenues for understanding the dynamics of learning in complex models.

While our method effectively characterizes fractal behaviors, it may not fully capture the complexity inherent in the trainability boundary. We computed the box dimension of these boundaries as a more feasible alternative to direct, uniform sampling from the trainability boundary set, which remains impractical. However, the box dimension is not without its limitations; for example, it is known that all rational numbers between 0 and 1 technically have a box dimension of $1$ \cite{strogatz2018nonlinear, falconer2007fractal}. Consequently, while the relative magnitudes of our computed box dimensions can be informative in assessing the degree of complexity, the absolute values themselves may not be entirely reliable.

Beyond mathematical limitations, constraints in our numerical implementation also impact the accuracy of the fractal dimensions we obtained. For instance, computational resources cap the largest feasible $N$, limiting the number of data points available for accurately fitting the fractal dimension. If a fractal boundary is densely packed within a very narrow range, a significantly large $N$ is required to discern its fractal nature, potentially causing us to overlook certain fractal behaviors when $N$ is limited. Interestingly, the practical significance of these fractal boundaries also comes into question; narrowly distributed boundaries are unlikely to be encountered in most applications, thus posing minimal risk. This observation led to a new insight: fractal dimension alone may not suffice to assess the risk posed by fractal boundaries. It also becomes essential to understand the distribution breadth of these boundary points. In our experiments, the maximum $N$ tested did not vary widely, suggesting that we may have consistently overlooked very narrow fractal boundaries. However, this might not be detrimental, as such boundaries are less likely to impact practical applications.

Our renormalization analysis, while effective in identifying roughness as a key parameter, exhibits limited generalizability. This analysis is restricted to simple functions with explicitly defined parameters, making our conclusions highly specific to the cases studied. Additionally, we cannot answer with this analysis why non-convexity of the loss function leads to the emergence of fractal behaviors. My initial concept was to establish a mapping that links different loss landscapes and learning rates, thereby preserving unchanged trainability. This mapping would ideally define an updating flow for function parameters and the learning rate. If successful, we could potentially transform the question of trainability into an investigation of where this updating flow stabilizes, using familiar functions such as the quadratic function as endpoints. However, the renormalization flow falls short in achieving this, as it cannot eliminate the perturbations. The first time I viewed the figures in \cite{sohldickstein2024boundary}, they reminded me of images of \href{https://science.nasa.gov/jupiter/}{Jupiter}, whose fractal-like surface arises from some fluid dynamics. This analogy suggests a future possibility where we might develop an updating flow for hyperparameters that mirrors principles from fluid dynamics.

In conclusion, substantial future research is necessary to more accurately capture the fractal behaviors of trainability boundaries. As we have discussed, developing a theory that can predict both the critical emergence of fractal trainability boundaries and their fractal dimensions is essential. Moreover, establishing connections to realistic loss functions from contemporary machine learning models is needed, particularly finding methods to characterize roughness in general loss functions lacking simple explicit formulas. Further exploration into the mechanisms that contribute to rough non-convexity is also required. With a deeper understanding of these phenomena, we could potentially develop strategies associated to model construction, dataset management, and optimizers that mitigate the risks associated with dangerous fractal trainability boundaries. By continuing to build on this foundation, we pave the way for more robust and predictable machine learning methodologies.

\section*{Methods}

We conducted large scale numerical experiments with Julia 1.8.4 on CPUs or with Python 3.9 on GPUs of MIT Supercloud \cite{reutherInteractiveSupercomputing402018}. The results have no notable difference. 
We ran small scale tests and analyze data with Python 3.10.9 on a laptop. All codes are available \href{https://github.com/liuyz0/FractalBoundary}{online}. In practice, we set number of segments $N = 2^n$ for integer $n$, and ran tests on $N_{\rm max} +1 = 2^{n_{\rm max}}+1$ learning rates evenly distributed in $[0,1.5]$.
Given hyperparameters and the loss function, we ran GD for 1000 steps, and classify bounded or divergent training based on the sum of loss values of the 1000 steps. We also classified bounded or divergent training based on whether GD cannot or can hit an upper bound. The latter classification may mistake some cases where GD first diverge but then converge. However, in the tests reported in the main text, the two classification methods do not have notable difference.
When analyzing data, we can choose $2^n+1$ ($n\leq n_{\rm max}$) evenly spaced points from the $2^{n_{\rm max}}+1$ points to analyze boundary segments at a coarse-grained level, which can give $|B_N|$ with $N = 2^n$. The largest $n_{\rm max}$ we tested is $32$. Most times, $n_{\rm max}=20$ is sufficient to yield a good fitting of fractal dimensions. We ran all numerical tests with data type \texttt{float64}, which is accurate enough for our choices of $n_{\rm max}$. 

The choice of learning rate range $[0,1.5]$ tested relies on the facts $f_0 = x^2$ has one trainability boundary at $s=1.0$ and we observe a lot of trainability boundaries when $s<1.0$ in practice (Fig.~\ref{fig:1}d). We prove  $f_0 = x^2$ has one trainability boundary as follows. By the definition of GD, on $f_0$, we have
\begin{equation}
x^{(k+1)} = x^{(k)} - 2sx^{(k)} = (1 - 2s)x^{(k)}.
\end{equation}
Convergence requires $|1 - 2s|<1$, which gives $0<s<1$ and thus completes the proof. We applied our numerical method to $f_0$ for a rational check and found indeed there is no fractal trainability boundary for $f_0$ (SI).

Details of the renormalization procedure are given as follows. For both additive and multiplicative perturbation cases, we can write the loss function in a form
\begin{equation}
f(x) = x^2 + \phi.
\end{equation}
By substituting $\tilde{x} = x/b$ and $\tilde{f}(\tilde{x}) = f(x) / \zeta$ into the original GD, we have
\begin{equation}
\tilde{x}^{(k+1)} = \tilde{x}^{(k)} - \tilde{s} \nabla \tilde{f}(\tilde{x}^{(k)}),
\end{equation}
with $\tilde{s} = s\zeta /b^2$ and
\begin{equation}
\tilde{f}(\tilde{x}) = \frac{b^2}{\zeta}\tilde{x}^2 + \frac{1}{\zeta} \phi(b\tilde{x}).
\end{equation}
Since we want the new function $\tilde{f}$ to have the same function form as $f$, we need the pre-factor of $\tilde{x}^2$, i.e., $b^2/\zeta$, to be one.
Consequently, we have $\zeta = b^2$ and the learning rate $\tilde{s} = s$ unchanged.
And for the additive perturbation case, where $\phi = \epsilon \cos(2\pi x/\lambda)$, if we want to write the transformed $\tilde{\phi} =\frac{1}{\zeta} \phi(b\tilde{x}) = \tilde{\epsilon} \cos(2\pi \tilde{x}/\tilde{\lambda})$, we will arrive the results $\tilde{\epsilon} = \epsilon/b^2$ and $\tilde{\lambda} = \lambda/b$.
Similarly, for the multiplicative perturbation case, since $\phi = x^2 \epsilon \cos(2\pi x/\lambda)$ and $\tilde{\phi} = \frac{1}{\zeta} \phi(b\tilde{x}) = \tilde{x}^2 \tilde{\epsilon} \cos(2\pi \tilde{x}/\tilde{\lambda})$, we will have $\tilde{\epsilon} = \epsilon$ and $\tilde{\lambda} = \lambda/b$.
Since $\tilde{x} = x/b$ is a one-to-one mapping, we know that changing the set of conditions $\{s, \epsilon, \lambda, x^{(0)}\}$ to $\{s, \tilde{\epsilon}, \tilde{\lambda}, x^{(0)}/b\}$ will only yield a rescaled GD trajectory but not change whether the trajectory diverge or not. In other words, the conditions $\{ \epsilon, \lambda, x^{(0)}\}$ have the same trainability boundary as $\{\tilde{\epsilon}, \tilde{\lambda}, x^{(0)}/b\}$.

\section*{acknowledgments}
It is a pleasure to thank Weijie Su for introducing this problem and highlighting the importance, Ziming Liu for discussions on the intuitions, J\"orn Dunkel for valuable discussions on aspects to explore, Jeff Gore for the support and inspiring discussions, and Jascha Sohl-Dickstein for valuable suggestions.



%


\setcounter{equation}{0}
\setcounter{figure}{0}
\setcounter{table}{0}
\setcounter{section}{0}
\makeatletter
\renewcommand{\theequation}{S\arabic{equation}}
\renewcommand{\thefigure}{S\arabic{figure}}
\renewcommand{\thesection}{S-\Roman{section}}


\section*{Supplementary Information}
All the raw data and comparison between fitted line and raw data points are available \href{https://github.com/liuyz0/FractalBoundary}{online}.

\begin{figure}
    \centering
    \includegraphics{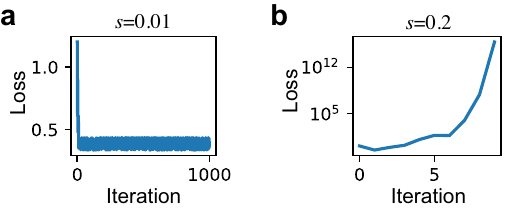}
    \caption{Training loss can be bounded or divergent. (a) An example obtained based on the multiplicative noise case with amplitude $\epsilon = 0.2$, wavelength $\lambda=0.1$, and learning rate $s = 0.01$, where the loss will decay and be bounded. (b) An example of divergent training based on the multiplicative noise case with amplitude $\epsilon = 0.2$, wavelength $\lambda=0.1$, and learning rate $s = 0.2$.}
    \label{fig:SIdyn}
\end{figure}

\begin{figure}
    \centering
    \includegraphics{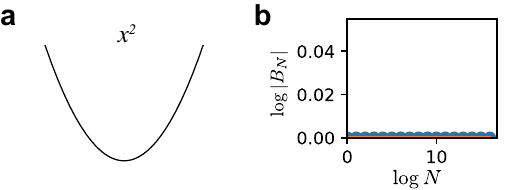}
    \caption{Sanity check on the quadratic loss function indicates our numerical method is not wrong. (a) The quadratic function $f_0=x^2$, on which we know there is only one trainability boundary $s=0$. (b) We found with our numerical method $|B_N|$ is always $1$, consistent with the theory.}
    \label{fig:SIsanity}
\end{figure}

We show two examples of training loss updating on our constructed loss landscapes in Fig.~\ref{fig:SIdyn}.
The sanity check of our method on the quadratic function $f_0 = x^2$ is presented in Fig.~\ref{fig:SIsanity}.

\begin{figure}
    \centering
    \includegraphics{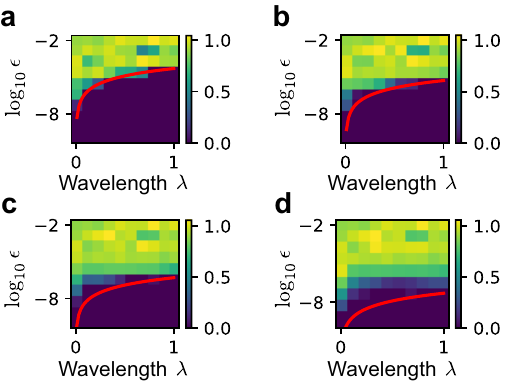}
    \caption{Numerical artifact supports that fractal behaviors emerge when the landscape is non-convex. When we increase the loss upper bound for classifying bounded and divergent training, GD can run on greater regions. On different parameter ($x$) regions, the function parameters $(\epsilon,\lambda)$ for multiplicative case ($f_\times (x)$) to be non-convex are different (red curves in the figure is boundary of convexity and non-convexity). Changing the upper bound from (a) \texttt{1e+3}, (b) \texttt{1e+4}, (c) \texttt{1e+5}, to (d) \texttt{1e+6}, the fractal behaviors all show up after the region GD runs on becomes non-convex (above the red curves).}
    \label{fig:SIartifact}
\end{figure}

The multiplicative case is always non-convex and should always display fractal trainability boundary. But if we study the second derivative,
\begin{align}
    \nabla^2 f_\times (x) = & 2 + 2\epsilon\cos\left(\frac{2\pi x}{\lambda}\right) - 4 \epsilon \frac{2\pi x}{\lambda} \sin\left(\frac{2\pi x}{\lambda}\right) \nonumber\\
    &- \epsilon \left(\frac{2\pi x}{\lambda}\right)^2 \cos\left(\frac{2\pi x}{\lambda}\right),
\end{align}
we find that $\nabla^2 f_\times (x)>0$ in a region $|x|$ is small enough. If we set an upper bound $f_{\rm max}$ to tell whether training is classified to be bounded or divergent and do not care the dynamics once it goes beyond $f_{\rm max}$, GD actually only sees the loss within the region $|x| < \sqrt{f_{\rm max}}$ roughly. And if $\epsilon$ is too small, in the region $|x| < \sqrt{f_{\rm max}}$, the loss may be convex and we may not see any fractal behaviors, which is a numerical artifact. To test the idea, we set different $f_{\rm max}$ values and stop GD and regard it as divergent once the loss reaches $f_{\rm max}$. We found too small $\epsilon$ indeed will make fractal behaviors vanish and the transition to fractal behaviors differs for different upper bounds $f_{\rm max}$ (Fig.~\ref{fig:SIartifact}). We next analyze when the loss within $|x| < \sqrt{f_{\rm max}}$ can be non-convex and see if this case corresponds to the emergence of fractal behaviors. The critical situation is that $\nabla^2 f_\times$ becomes zero near $|x| = \sqrt{f_{\rm max}}$. Since $\sqrt{f_{\rm max}}$ is very large, the quadratic term dominants among terms having $\epsilon$, we have the minimum second derivative near $|x| = \sqrt{f_{\rm max}}$ being
\begin{align}
    \min \nabla^2 f_\times \approx  2 - \epsilon f_{\rm max} \left(\frac{2\pi}{\lambda}\right)^2.
\end{align}
By setting the above estimation to be zero, we reach the boundary of non-convexity for loss within $|x| < \sqrt{f_{\rm max}}$ as
\begin{equation}
    \epsilon = \frac{1}{f_{\rm max}} \frac{\lambda^2}{2\pi^2}.
\end{equation}
These estimated boundaries are plotted in the $(\epsilon,\lambda)$ plane for different $f_{\rm max}$ values (red curves in Fig.~\ref{fig:SIartifact}). Note greater $\epsilon$ means non-convexity, we found it is true that fractal behaviors only show up when the part of loss function GD can see becomes non-convex (non-zero fractal dimensions are all above the red curves). We also note that with increasing upper bound $f_{\rm max}$, the boundary of non-convexity (red curves) tend to be smaller and smaller than the boundary of fractal behaviors. This might due to the fact the region $|x|<\sqrt{f_{\rm max}}$ is expanding with larger $f_{\rm max}$ and near critical $(\epsilon,\lambda)$ for non-convexity, it is more difficult to see the non-convex part near $|x|\approx \sqrt{f_{\rm max}}$. In conclusion, the numerical artifact help to further prove the idea that loss becoming non-convex leads to the emergence of fractal behaviors in training.

\begin{figure*}
    \centering
    \includegraphics[width = \textwidth]{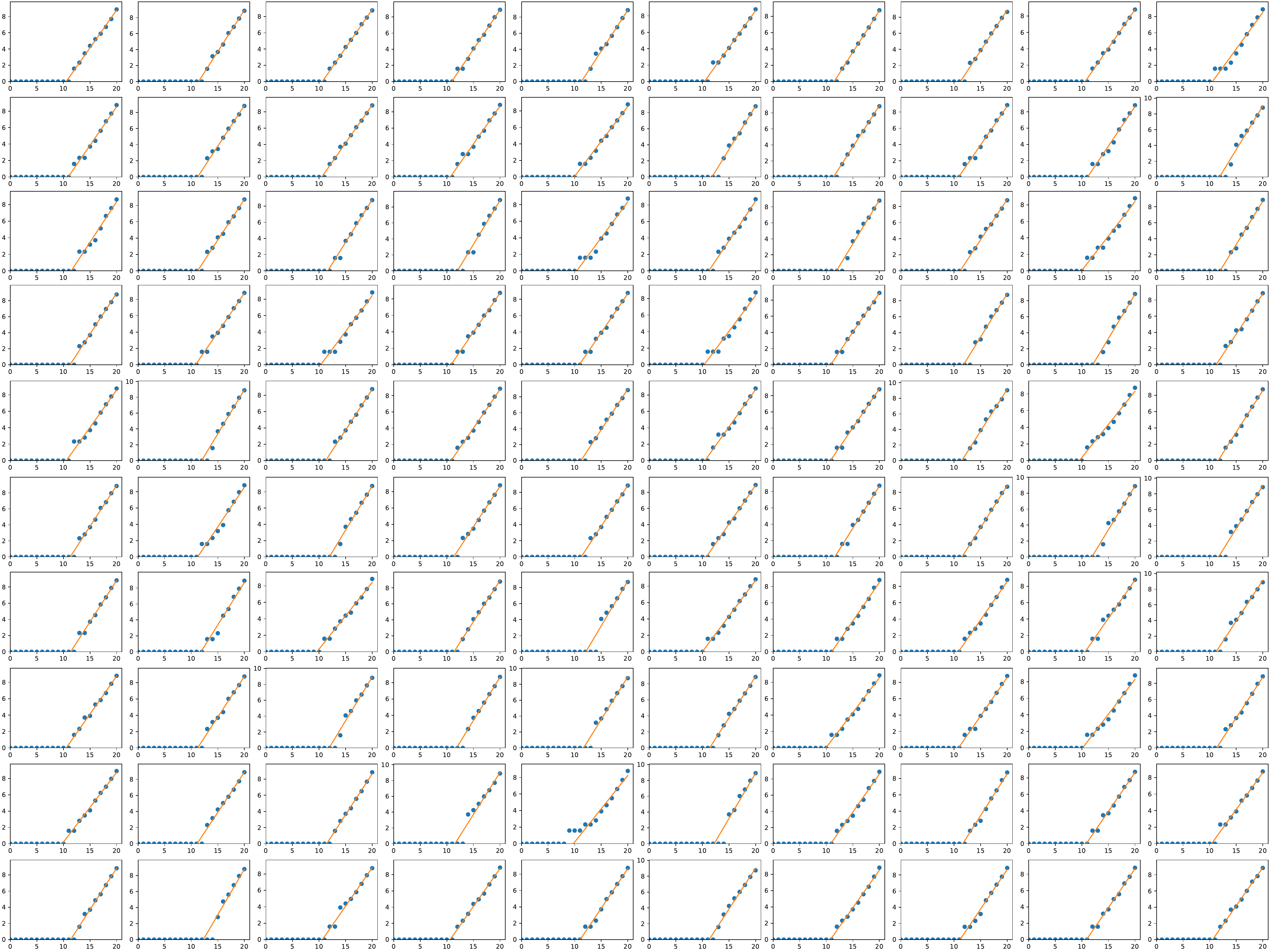}
    \caption{Tests suggest that initial parameter $x^{(0)}$ may not affect fractal dimension. We set for the additive noise case $\epsilon = 0.2$ and $\lambda=0.1$, and sampled 100 different initial conditions $x^{(0)}$ uniformly from $[-5,5]$. The fractal dimension averaged over initial conditions is $0.98$ with a standard deviation $0.08$.}
    \label{fig:7si}
\end{figure*}

\begin{figure*}
    \centering
    \includegraphics[width = \textwidth]{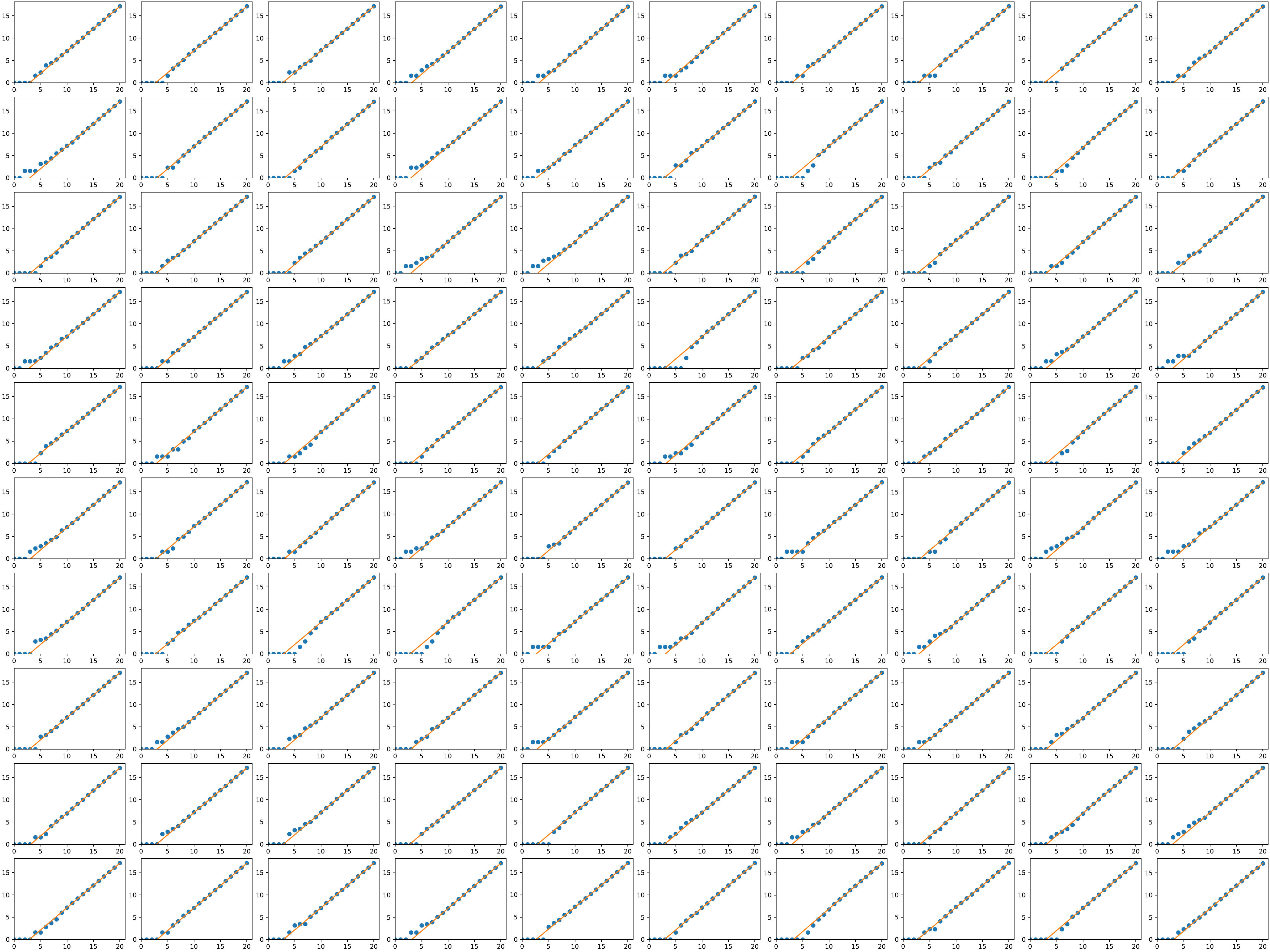}
    \caption{Tests suggest that initial parameter $x^{(0)}$ may not affect fractal dimension. We set for the multiplicative noise case $\epsilon = 0.2$ and $\lambda=0.1$, and sampled 100 different initial conditions $x^{(0)}$ uniformly from $[-5,5]$. The fractal dimension averaged over initial conditions is $1.00$ with a standard deviation $0.01$.}
    \label{fig:6si}
\end{figure*}

We checked the dependence of fractal dimension on the initial condition of parameter $x^{(0)}$ in Fig.~\ref{fig:7si} and \ref{fig:6si}, which suggest initial parameter $x^{(0)}$ may not affect fractal dimension.


\end{document}